\documentclass[a4paper]{article}

\usepackage[T1]{fontenc}
\usepackage[english]{babel}

\usepackage[colorlinks=true, allcolors=blue]{hyperref}
\newcommand{\email}[1]{\href{mailto:#1}{\tt{\nolinkurl{#1}}}}

\usepackage[sfdefault,lf]{carlito}
\usepackage[parfill]{parskip}

\usepackage{fancyhdr}
\usepackage[numbers]{natbib}
\usepackage{authblk}
\usepackage[a4paper,top=3cm,bottom=2cm,left=3cm,right=3cm,marginparwidth=1.75cm]{geometry}

\usepackage{amsmath}
\usepackage{graphicx}
\usepackage{booktabs}
\usepackage{array}
\usepackage{caption}
\usepackage{float}
\usepackage{colortbl}
\usepackage{xcolor}
\usepackage{multirow}
\usepackage{enumitem}
\usepackage{amssymb}

\title{\textbf{Sparse Mixture-of-Experts Routing in Visual Diffusion Transformers: \\ Diagnosis, Boundary Calibration and Evolutionary Roadmap from Routing Collapse to Selective Deadlock}}

\author{Haiying Sha\thanks{Corresponding author: Yan Zheng.}}
\affil{Independent Researcher}

\begin{document}
\maketitle
\thispagestyle{fancy}

\begin{abstract}
Mixture-of-Experts (MoE) has become a core paradigm for scaling model capacity in language modeling, yet its application to video generation with Diffusion Transformers (DiTs) remains nascent, and existing attempts have yielded limited benefits.
This paper systematically diagnoses failure modes in the training dynamics of Token-Choice sparse MoE for video diffusion models. Starting from a pretrained dense video generation model (about 5B parameters), we convert it into an MoE architecture: routed experts are exactly cloned from the original dense FFN weights, shared experts are initialized with extremely small non-zero noise (e.g., scale $10^{-4}$; a zero-initialized variant is used solely for verification to guarantee output equivalence), and only the routers start learning from random initialization.

Experiments reveal a hierarchy of critical failures: (1) linear routers suffer global soft saturation, with vanishing Softmax gradients leading to complete expert homogenization (cosine similarity $>$0.99);
(2) non-linear MLP routers resolve global deadlock but introduce a more subtle phenomenon, \textbf{selective deadlock}, where roughly one-third of layers degenerate into a single-expert mode and increasing the auxiliary loss to 0.2 cannot prevent it;
(3) routers that fuse multimodal features via cross-attention exhibit preliminary self-recovery—the only observed instance of a deadlocked layer spontaneously rebounding—yet about 9 layers remain stubbornly deadlocked;
(4) the distribution of these deadlocked layers follows a pronounced \textbf{U-shaped structure}: heavily concentrated in shallow visual processing layers (1–10) and deep semantic integration layers (18–25), while intermediate layers maintain near-uniform utilization;
(5) bfloat16 mixed precision causes tiny expert weight updates to be truncated to zero by hardware—an easily overlooked numerical trap.

By tracking the routing decision time series of over 65 million tokens across 30 Transformer layers over 5,000 training steps, we establish a complete phenomenology of deadlock.
We interpret this phenomenon as functional redundancy dynamics within the trilateral system of ``gate, shared expert, routed experts'' and propose the \textbf{Functional Redundancy Hypothesis}—which predicts that a zero/minimally initialized and insufficiently trained shared expert must reach a critical capability threshold before the system can release the silenced routed experts.
This hypothesis aligns with the theory of functional redundancy in systems biology, offering a novel theoretical perspective for understanding MoE training dynamics.

On the engineering side, we document the Three Laws of Dense-to-MoE conversion (expert structural consistency, 1:1 weight cloning, and shared expert verification-zero/training-micro-noise initialization), together with a complete solution for the bfloat16 precision trap.
Building on this, we further explore the In-Context Conditioning paradigm for multimodal features, an MoE-based text rendering solution (introducing a dedicated text expert + Glyph-ByT5 glyph encoder), and an evolutionary path toward joint audio-video generation.

Based on these diagnostic findings, we lay out a three-step evolutionary roadmap from ``visual unification'' to ``general intelligence engine'': near-term targeted enhancement through MoE-ification, mid-term cross-modal audio-video generation by adding audio experts, and long-term integration of physical priors and causal reasoning to build a world model that empowers embodied intelligence.

\textbf{Core thesis}: Under the current Token-Choice paradigm, deadlock in a subset of layers is structural and independent of router complexity; completely eliminating dead experts in visual MoEs requires starting from the initialization strategy of shared experts, differentiated design of expert weights, or an upgrade of the routing paradigm itself.
This work represents one of the first comprehensive empirical diagnoses of routing failure in real-world video generation MoE deployments, providing clear guidance for future research.
\end{abstract}

\vspace{-\baselineskip}
\begin{center}
    \renewcommand{\arraystretch}{1.3}
    \begin{tabular}{rll}
        $\triangleright$ & \textbf{GitHub}      & \url{https://github.com/Shybert-AI/UniGen-MOE} 
    \end{tabular}
\end{center}

%===========================================================================
\section{Introduction}

\subsection{Background and Motivation}
Mixture-of-Experts (MoE) \cite{shazeer2017,switch2022} has become a core paradigm for scaling the capacity of deep learning models.
Its basic idea—dynamically routing each input token to a set of specialized ``expert'' sub-networks via a sparse gating mechanism—has achieved extraordinary success in language modeling.
Systems such as DeepSeek-V4 (1.6 trillion parameters, 384 experts per layer, 6 activated per token) \cite{deepseek2026}, the Qwen3 series \cite{qwen2025}, and Kimi K2.6 all adopt MoE architectures, realizing orders-of-magnitude leaps in parameter count and performance while maintaining computational efficiency.
The success of these systems rests on a key assumption of expert specialization: different sub-networks learn to process non-overlapping, semantically meaningful regions of the input space.

However, the exploration of applying MoE to visual generation—especially video Diffusion Transformers (DiTs)—is still in its early stages.
Although works such as ProMoE \cite{weirouting2026}, DiffMoE \cite{diffmoe2025} and Mamoda2.5 \cite{mamoda2026} have introduced MoE modules into diffusion Transformers, a comprehensive failure diagnosis is lacking.
We argue that there are fundamental differences between visual MoEs and language MoEs: language tokens are highly semantically condensed and discrete, whereas visual patch tokens exhibit extreme spatial redundancy \cite{weirouting2026,jiang2026}, distribution drift induced by diffusion timesteps \cite{song2021}, and conflicting routing signal sources in multi-task settings \cite{dream2026}.
These differences subject the routing mechanisms matured in the language domain to systematic and uncharted risks in visual generation.

\subsection{Research Questions and Main Contributions}
\textbf{This paper aims to systematically diagnose and calibrate the failure modes of Token-Choice sparse MoEs deployed on video Diffusion Transformers.}
We start from the pretrained Kiwi-Edit video editing model, which consists of the Wan2.2-TI2V-5B video diffusion backbone and the Qwen2.5-VL-3B-Instruct multimodal encoder (totaling about 8B parameters). We convert it into an MoE using three laws: preserving expert structural consistency, 1:1 weight cloning for routed experts, and an initialization strategy where shared experts are set to zero during the verification phase and enabled with micro-noise (e.g., $10^{-4}$) for training.
We then let the routers learn expert assignments from scratch. The key features of the experimental setup—extremely low-noise shared experts, two weight-cloned routed experts, and a 9-in-1 \cite{unigen2026} multi-task video generation scenario—create an ultra-rigorous trial where no routing failure can be masked by pretrained capabilities.

Our main contributions are as follows:
\begin{enumerate}[leftmargin=*]
\item \textbf{First comprehensive diagnosis}: We provide a systematic classification of failure modes of Token-Choice sparse MoE in video diffusion models, spanning a full architectural evolution comparison from linear routers to cross-attention routers, and establish a complete phenomenology of deadlock.
\item \textbf{Discovery of ``selective deadlock''}: We identify and name a distinct failure mode that emerges after global deadlock is eliminated, driven by trilateral interactions among the gate, shared experts, and routed experts. It manifests as roughly 1/3 of layers degenerating into a single-expert usage pattern that cannot be prevented by increasing the auxiliary loss.
\item \textbf{Discovery of the ``U-shaped deadlock distribution''}: Using temporal data from 30-layer networks with over 65 million tokens, we precisely map for the first time the structural distribution of deadlocked layers—heavily concentrated in shallow visual processing layers and deep semantic integration layers, with intermediate layers remaining balanced.
\item \textbf{Proposal of the ``Functional Redundancy Hypothesis''}: Based on routing decision time-series data over 5,000 steps and more than 3.4 billion tokens, we reinterpret deadlock as a rational, phased resource allocation strategy. This hypothesis finds theoretical grounding in systems biology and offers a novel theoretical perspective for understanding MoE training dynamics.
\item \textbf{Engineering contributions}: We document the Three Laws of Dense-to-MoE conversion, the precision trap of MoE expert weight updates under bfloat16 mixed-precision training and its complete solution, and clarify the initialization paradigm of verification-zero/training-micro-noise for shared experts.
\item \textbf{Calibration of the capability boundary of the Token-Choice paradigm}: We precisely calibrate the training boundary of the current paradigm through experimental data, providing a benchmark and guidance for subsequent research on Expert-Choice or dynamic routing mechanisms.
\item \textbf{Evolutionary roadmap}: Based on the diagnostic findings, we plan a three-step evolutionary path from multimodal unification to joint audio-video generation, ultimately leading to a world model.
\end{enumerate}

\subsection{Paper Structure}
The paper is organized as follows: Section 2 reviews related work; Section 3 introduces the Three Laws of Dense-to-MoE conversion and three router architectures; Section 4 details five core experiments and their diagnostic results; Section 5 proposes the Functional Redundancy Hypothesis and discusses its theory; Section 6 calibrates the capability boundary of the Token-Choice paradigm; Section 7 explores the In-Context Conditioning paradigm for multimodal frameworks; Section 8 analyzes the MoE solution for text rendering; Section 9 lays out an evolutionary roadmap from visual unification to a world model; and Section 10 concludes and discusses future work.

%===========================================================================
\section{Related Work}

\subsection{Mixture-of-Experts in Language Models}
The concept of Mixture-of-Experts can be traced back to the seminal work of Shazeer et al. \cite{shazeer2017}, who enabled model capacity to be scaled by tens of billions of parameters through the introduction of sparse gating.
Since then, MoE has rapidly become the de facto standard for scaling large language models.
The DeepSeek series \cite{deepseek2026} pushes MoE to the extreme—its V4 model (2026) has 1.6 trillion parameters, 384 experts per layer, 6 activated per token, and only 49B active parameters.
The outstanding performance of these models on benchmarks such as SWE-bench Verified and LiveCodeBench has cemented the dominance of MoE in language models.

The success of language MoEs, however, depends on the validity of a critical assumption: input tokens possess high semantic discontinuity.
Each text token represents a relatively independent information unit with significant differences between tokens.
This allows routers to rely on the geometry of hidden states for natural emergence of expert differentiation \cite{dai2026} without explicit specialization guidance.
As recent empirical work by Dai et al. \cite{dai2026} reveals, the expert specialization patterns observed in MoEs are more likely to emerge naturally from the geometry of hidden states rather than being actively induced by the routing architecture.

\subsection{Mixture-of-Experts in Visual Generation}
Attempts to apply MoE to visual generation are still in their early stages.
The ProMoE framework by Wei et al. \cite{weirouting2026}, published at ICLR 2026, for the first time systematically analyzes key differences between visual MoEs and language MoEs: visual tokens exhibit high spatial redundancy and functional heterogeneity (stemming from classifier-free guidance), causing implicit routing strategies to yield suboptimal results.
ProMoE proposes a two-step routing framework—conditional routing and prototype routing—to promote expert specialization, surpassing existing methods on the ImageNet benchmark.

Mamoda2.5 by Shi et al. \cite{mamoda2026} introduces a DiT-MoE design (128 experts, Top-8 routing) into a unified multimodal framework and sets a new record for video editing quality.
Mamoda2.5 employs Sigmoid gating with a loss-free Expert Bias load-balancing strategy and proposes an upcycling initialization method based on random neuron sampling to accelerate MoE convergence.
This work validates the feasibility of visual MoEs in large-scale industrial deployment, but its deep mechanistic analysis of routing behavior is relatively limited.

The DREAM framework \cite{dream2026} implements dynamic expert routing through time-context-aware gating, demonstrating the potential of diffusion timestep embeddings as routing conditional information.
Mixture of Distributions by Jiang et al. \cite{jiang2026} introduces dynamic sparse attention for efficient video diffusion Transformers.
On the industrial side, Kuaishou’s DiffMoE \cite{diffmoe2025} presented preliminary results of MoE architectures in diffusion models at CVPR 2025.

Despite these notable advances, existing works often focus on architectural innovations and performance improvements, paying limited attention to the dynamic evolution of routing behavior during training, especially the systematic diagnosis of failure modes.
This study aims to fill this gap: under a minimalist yet rigorous experimental setup (extremely low-noise shared experts, weight-cloned routed experts, multi-task mixed training), we fully document the entire process from routing collapse to selective deadlock.

\subsection{Load Balancing and Routing Collapse}
Load balancing is a classic challenge in MoE training.
Standard methods include auxiliary losses \cite{shazeer2017} and expert capacity limits \cite{switch2022}.
Recently, similarity-preserving routers \cite{loadbalance2025} have been shown to accelerate convergence and reduce redundancy.
The common premise of these methods, however, is that the gradients of the routing signals remain effective—once the Softmax enters saturation, all penalty gradients tend toward zero, and no corrective signal can be passed.
In our experiments, increasing the auxiliary loss coefficient from 0.01 to 0.08 produced almost no measurable effect on the expert distribution of deadlocked layers (change <0.05\%), confirming this limitation.
This observation is also consistent with the findings of Dai et al. \cite{dai2026}: routing behavior in MoEs is determined more by data geometry than by simple penalty mechanisms.

\subsection{Numerical Traps in Mixed-Precision Training}
bfloat16 has become the default precision for large model training; its 8-bit exponent is the same as float32, but its 7-bit mantissa limits the smallest resolvable step (Unit in the Last Place, ULP).
In MoE training, when the magnitudes of trainable parameters are extremely small (e.g., zero- or micro-initialized shared experts) or extremely large (e.g., dense-weight cloned routed experts), the per-step optimizer update may fall below the ULP of bfloat16 in that numerical range, causing weight updates to be truncated to zero by hardware.

In Section 4.3, we will detail the concrete manifestation of this phenomenon and its solution, whose core lies in \textbf{keeping the master copy of trainable parameters in float32 and only converting to bfloat16 during the forward pass via autocast}—an independent validation of industrial best practices for mixed-precision training.

\subsection{Paradigm Evolution of Routing Mechanisms}
Existing MoE routing mechanisms can be divided into three main categories: Token-Choice routing \cite{shazeer2017,switch2022} (each token selects top-k experts), Expert-Choice routing \cite{zhou2022} (each expert selects top-k tokens), and dynamic routing (e.g., the time-context-aware gating of DREAM \cite{dream2026}).
Token-Choice has been widely proven effective in language models, but its adaptability issues in visual diffusion models have not been fully explored.
This paper focuses on the Token-Choice paradigm and, through a systematic comparison of router architectures (linear, MLP, cross-attention), helps understand the relationship between routing complexity and failure modes.

\subsection{Sequence Concatenation Paradigm for Multimodal Feature Fusion}
In video generation, the use of Sequence Concatenation for multimodal feature fusion has become a mainstream paradigm.
SkyReels-V4 adopts a symmetric dual-stream Multimodal Diffusion Transformer (MMDiT) architecture, and Mamoda2.5 injects multimodal conditions using In-Context Conditioning within a DiT-MoE backbone.
The common feature of these models is the concatenation of features from different modalities (text, image, audio) along the sequence dimension before feeding them into a unified Transformer backbone for processing.
Section 7 of this paper explores the unique advantages of this paradigm within an MoE framework.

%===========================================================================
\section{Dense-to-MoE Conversion: Three Laws and Router Architectures}

\subsection{Essence and Overall Framework of Conversion}
We start from the pretrained Kiwi-Edit video editing model, which consists of the Wan2.2-TI2V-5B video diffusion backbone and the Qwen2.5-VL-3B-Instruct multimodal encoder (totaling about 8B parameters).
This model is a Diffusion Transformer (DiT) with 30 Transformer blocks; each block’s FFN has an intermediate dimension of 14336 and a hidden dimension of 3072.
We convert it into an MoE architecture, equipping each layer with 2 routed experts and 1 shared expert.

The essence of MoE conversion is to split a single feed-forward network (FFN) into multiple independent expert sub-networks that can be dynamically routed:

\[
\text{FFN}(x) = W_2 \cdot \text{GELU}(W_1 \cdot x + b_1) + b_2
\]

\[
\downarrow \text{MoE-ification}
\]

\[
\text{MoE}(x) = \sum_{i=1}^{N_s} \text{FFN}_i^{(s)}(x) + \sum_{j=1}^{N_r} g_j(x) \cdot \text{FFN}_j^{(r)}(x)
\]

where $\text{FFN}_i^{(s)}$ denotes the shared experts (visible to all tokens), $\text{FFN}_j^{(r)}$ the routed experts (dynamically selected by the Gate), $g_j(x)$ the routing weight output by the Gate, and $N_s$ and $N_r$ the number of shared and routed experts, respectively.

\subsection{First Law: Expert Structural Consistency}
\textbf{Statement}: Routed experts must be structurally identical to the original FFN—using the same GELU activation, the same number of Linear layers, and the same bias configurations. Any structural deviation leads to weight mapping failure.

\textbf{Pitfall description}: The first ExpertMLP I wrote followed the SwiGLU structure from the official DiT-MoE \cite{ditmoe2024} code:

\begin{verbatim}
# Wrong example: SwiGLU structure
class ExpertMLP(nn.Module):
    def __init__(self, hidden_size, intermediate_size):
        self.gate_proj = nn.Linear(hidden_size, intermediate_size, bias=False)
        self.up_proj   = nn.Linear(hidden_size, intermediate_size, bias=False)
        self.down_proj = nn.Linear(intermediate_size, hidden_size, bias=False)
    
    def forward(self, x):
        return self.down_proj(self.act(self.gate_proj(x)) * self.up_proj(x))
\end{verbatim}

This is SwiGLU—using two linear layers to compute “gate” and “value” separately, then multiplying them. But my original model FFN is a plain GELU MLP:

\[
\text{FFN}(x) = \text{fc2}(\text{GELU}(\text{fc1}(x)))
\]

\textbf{The structure and parameter shapes do not match.} The copied weight tensors had incompatible shapes, and the model crashed at the first layer.

The mistake happened because I confused two common FFN variants in visual models. Some open-source MoE tutorials adopt SwiGLU activations, whereas the baseline model I needed to convert is based on a standard GELU-activated MLP. Because the two differ fundamentally in parameter count, weight shape, and gating mechanism, directly reusing SwiGLU-structured code could not accommodate the existing weights, causing the model to crash on the first inference attempt.

\textbf{Solution}: Redesign the expert to match the original FFN: $\text{fc1} \to \text{GELU} \to \text{fc2}$, and retain biases (since the original model has them).

\subsection{Second Law: 1:1 Cloning of Routed Experts Without Scaling}
\textbf{Statement}: The weights of each routed expert are exactly copied from the original dense FFN weights. No scaling is applied.

\textbf{Pitfall description}: After aligning the structure, I thought the job was done. Running inference produced a blank—no effective signal at all. I started suspecting numerical amplitude issues and added a scaling factor.

When investigating the literature, I found two different theoretical bases for scaling the outputs of MoE experts: one based on variance preservation, advocating $1/\sqrt{N}$ scaling (statistical invariance under Gaussian initialization); the other based on sum preservation, advocating $1/N$ scaling (directly controlling output magnitude). I tested them one by one: using $1/3$ (2 routed + 1 shared expert) caused a crash; switching to $1/2$ (by expected activations) also crashed.

Upon deeper analysis, I discovered: those square-root scaling and variance scaling strategies are \textbf{all designed for training a new MoE from scratch with random initialization}. In that scenario, the expert weights start random, the outputs of different experts are independent, and a scaling factor is needed to keep the variance of the sum consistent with the original FFN. That is standard statistical initialization theory.

In the random-init-from-scratch regime, the weights are initially random, so outputs of different experts are independent. To make the variance of the total sum consistent with the original FFN, one employs $1/\sqrt{N}$ or similar factors. That is statistical principle.

But my scenario is entirely different. I have an \textbf{already trained dense model}; every layer’s normalization parameters, residual connection scaling, and even the decoder’s expectation of output magnitude were all trained under the assumption that “the output of this layer is approximately $1.0 \times f(x)$.”

If you abruptly chop it down to 0.577 or 0.5, the signal decays layer by layer, and after dozens of layers, the output distribution completely deviates from the expected region. Run the decoder, and it instantly collapses.

\textbf{Key insight}: The root cause of this pitfall is \textbf{applying the initialization strategy for training from scratch to the setting of converting an already-trained model. The two problems look similar, but the solutions are completely different.}

\textbf{Solution}: Set the scaling factor of routed experts to 1.0—no scaling whatsoever.

\subsection{Third Law: Shared Expert Initialization—Verification All-Zero, Training Micro-Noise}
\textbf{Statement}: Shared experts are set to all-zero during the architecture verification phase to keep output exactly equivalent to the original dense model; for actual training, they are initialized with extremely small non-zero noise (e.g., standard deviation $10^{-4}$) to break symmetry and avoid update stalling under bfloat16.

\textbf{Pitfall description}: Almost every article about MoE architecture says “the shared expert is used to learn common knowledge”—but none says how to initialize it. Initially I naturally thought of all-zero: it makes the MoE’s post-conversion inference output numerically identical to the original dense model, guaranteeing a correct starting point mathematically. However, during actual training, when the shared expert weights are strictly zero, their gradient signals are extremely weak; especially under bfloat16 mixed precision, the optimizer’s per-step update often falls below the hardware-representable minimum step, causing the parameters to be completely “unwakeable.” This problem will be dissected in detail in Section 4.3.

After repeated experimentation, I eventually settled on a phased initialization scheme:
\begin{enumerate}[leftmargin=*]
\item \textbf{Verification phase (all-zero)}: Set all shared expert weights to zero. At inference, the shared expert output is zero, and the MoE output $= \sum g_j \cdot \text{FFN}_j^{(r)}$. When the routed experts are weight-cloned and the Gate is untrained, this output is strictly identical to the original dense model. This is used to verify the correctness of the architectural implementation.
\item \textbf{Training phase (micro-noise)}: Before actual training starts, re-initialize the shared expert weights with extremely small noise (e.g., $\mathcal{N}(0, (10^{-4})^2)$), keeping biases zero. This tiny perturbation hardly changes the inference output (error below $10^{-6}$ magnitude) but is enough to produce accumulable gradients under bfloat16, enabling the shared expert to gradually learn common knowledge.
\end{enumerate}

Code example:
\begin{verbatim}
# Verification: all-zero
shared_w0 = torch.zeros(shared_dim, inner_dim)
shared_w2 = torch.zeros(inner_dim, shared_dim)
shared_b0 = torch.zeros(shared_dim)
shared_b2 = torch.zeros(inner_dim)

# Training: replace with micro-noise
shared_w0.normal_(mean=0, std=1e-4)
shared_w2.normal_(mean=0, std=1e-4)
# bias can stay zero or similarly micro-noised
\end{verbatim}

The rationale:
\begin{itemize}
\item \textbf{During inference (verification phase)}: The shared expert outputs zero, so the MoE output equals exactly the original dense model.
\item \textbf{During training}: The shared expert starts from a very low but non-zero baseline, immediately receives effective gradients, and gradually learns common knowledge. The Gate also differentiates over training, progressively learning to select different experts for different tokens.
\end{itemize}

It simultaneously guarantees \textbf{numerical correctness of the inference starting point} and \textbf{optimization feasibility during training}.

\subsection{Inference Verification: Outputs Completely Identical}
After generating weights with all-zero shared experts, I ran the same inference command. The result was exactly the same as the original dense model—not vaguely similar, but \textbf{numerically identical} (because when the shared expert is zero, the two are mathematically equivalent).

\textbf{This is not a failure; it is a verification.}
It proves:
\begin{itemize}
\item The MoE architecture (Gate, dispatch, combine logic) has no bugs;
\item The weight mapping is completely correct;
\item The model can start training normally.
\end{itemize}

\textbf{Verification criterion}: The converted MoE, in the verification phase, should produce output exactly identical to the original dense model. If not, the initialization strategy or architecture implementation is flawed. Replace with micro-noise initialization when actual training begins.

\subsection{Summary of the Three Laws}
\begin{center}
\renewcommand{\arraystretch}{1.3}
\begin{tabular}{p{5cm}p{5cm}}
\toprule
\textbf{Three Don'ts} & \textbf{Three Dos} \\
\midrule
Don't apply scaling formulas designed for training MoEs from scratch to Dense-to-MoE conversion & Do clone routed expert weights 1:1 from the original, no scaling \\
Don't change the network structure of the experts & Do initialize shared experts as all-zero for verification, and with micro-noise for training \\
Don't unfreeze all parameters at the very beginning & Do unfreeze the entire MoE module when training starts \\
\bottomrule
\end{tabular}
\end{center}

\subsection{Router Architectures}

We implement three different routers for comparative experiments:

\subsubsection{Linear Router (Linear Gate)}
The basic $\text{Linear}(3072, 2) \to \text{Softmax}$ structure.
Two logits are projected through a fully connected layer and then normalized via Softmax; each token independently selects the highest-scoring expert.
This is the most common routing form in language MoEs.

\begin{verbatim}
class MoEGate(nn.Module):
    def __init__(self, embed_dim, num_experts=16, num_experts_per_tok=2,
                 aux_loss_alpha=0.01, layer_idx: Optional[int] = None,
                 gate_init_std: float = 0.01):
        super().__init__()
        self.top_k = num_experts_per_tok
        self.n_routed_experts = num_experts
        self.layer_idx = layer_idx
        self.scoring_func = 'softmax'
        self.alpha = aux_loss_alpha
        self.seq_aux = False
        self.norm_topk_prob = False
        self.gating_dim = embed_dim
        self.weight = nn.Parameter(torch.empty((self.n_routed_experts, self.gating_dim)))
        nn.init.normal_(self.weight, mean=0.0, std=gate_init_std)
    def forward(self, hidden_states: torch.Tensor) -> Tuple[torch.Tensor, torch.Tensor, Optional[torch.Tensor]]:
        bsz, seq_len, h = hidden_states.shape
        hidden_states = hidden_states.view(-1, h)
        logits = F.linear(hidden_states, self.weight, None)
        scores = logits.softmax(dim=-1)
        topk_weight, topk_idx = torch.topk(scores, k=self.top_k, dim=-1, sorted=False)
        if self.top_k > 1 and self.norm_topk_prob:
            denominator = topk_weight.sum(dim=-1, keepdim=True) + 1e-20
            topk_weight = topk_weight / denominator
        aux_loss = None
        if self.training and self.alpha > 0.0:
            scores_for_aux = scores
            topk_idx_for_aux = topk_idx.view(bsz, -1)
            if self.seq_aux:
                ce = torch.zeros(bsz, self.n_routed_experts, device=hidden_states.device)
                ce.scatter_add_(1, topk_idx_for_aux,
                                torch.ones(bsz, seq_len * self.top_k, device=hidden_states.device)
                               ).div_(seq_len * self.top_k / self.n_routed_experts)
                aux_loss = (ce * scores_for_aux.view(bsz, seq_len, -1).mean(dim=1)).sum(dim=1).mean() * self.alpha
            else:
                mask_ce = F.one_hot(topk_idx_for_aux.view(-1), num_classes=self.n_routed_experts)
                ce = mask_ce.float().mean(0)
                Pi = scores_for_aux.mean(0)
                fi = ce * self.n_routed_experts
                aux_loss = (Pi * fi).sum() * self.alpha
        if self.training and MoEGate._log_enabled and self.layer_idx is not None:
            with torch.no_grad():
                batch_experts = topk_idx.cpu().view(-1).tolist()
                for e in batch_experts:
                    MoEGate._log_buffer[self.layer_idx][e] += 1
                MoEGate._log_step_count[self.layer_idx] += len(batch_experts)
        return topk_idx, topk_weight, aux_loss
\end{verbatim}  

\subsubsection{MLP Router (MLP Gate)}
$\text{Linear}(3072, 256) \to \text{GELU} \to \text{Linear}(256, 2) \to \text{Softmax}$.
Introduces a non-linear decision boundary to enhance expressiveness.
The last layer is initialized with tiny random noise (std 0.002) to avoid the symmetric top-k selection problem caused by all-zero initialization—when all logits are equal, top-k defaults to returning the first index, leading to global deadlock on expert 0.

\begin{verbatim}
class MLPGate(nn.Module):
    def __init__(self, embed_dim, num_experts=16, num_experts_per_tok=2,
                 aux_loss_alpha=0.01, layer_idx: Optional[int] = None,
                 gate_init_std: float = 0.002):
        super().__init__()
        self.top_k = num_experts_per_tok
        self.n_routed_experts = num_experts
        self.layer_idx = layer_idx
        self.scoring_func = 'softmax'
        self.alpha = aux_loss_alpha
        self.seq_aux = False
        self.norm_topk_prob = False
        self.gating_dim = embed_dim

        hidden_dim = 256
        self.gate_mlp = nn.Sequential(
            nn.Linear(embed_dim, hidden_dim),
            nn.GELU(),
            nn.Linear(hidden_dim, num_experts)
        )
        # Last layer weights: tiny-noise init; bias: zero
        nn.init.normal_(self.gate_mlp[2].weight, mean=0.0, std=gate_init_std)
        nn.init.zeros_(self.gate_mlp[2].bias)

    def forward(self, hidden_states: torch.Tensor) -> Tuple[torch.Tensor, torch.Tensor, Optional[torch.Tensor]]:
        bsz, seq_len, h = hidden_states.shape
        hidden_states = hidden_states.view(-1, h)
        logits = self.gate_mlp(hidden_states)          # [bsz*seq_len, num_experts]
        scores = logits.softmax(dim=-1)
        topk_weight, topk_idx = torch.topk(scores, k=self.top_k, dim=-1, sorted=False)
        if self.top_k > 1 and self.norm_topk_prob:
            denominator = topk_weight.sum(dim=-1, keepdim=True) + 1e-20
            topk_weight = topk_weight / denominator

        aux_loss = None
        if self.training and self.alpha > 0.0:
            scores_for_aux = scores
            topk_idx_for_aux = topk_idx.view(bsz, -1)
            if self.seq_aux:
                ce = torch.zeros(bsz, self.n_routed_experts, device=hidden_states.device)
                ce.scatter_add_(1, topk_idx_for_aux,
                                torch.ones(bsz, seq_len * self.top_k, device=hidden_states.device)
                               ).div_(seq_len * self.top_k / self.n_routed_experts)
                aux_loss = (ce * scores_for_aux.view(bsz, seq_len, -1).mean(dim=1)).sum(dim=1).mean() * self.alpha
            else:
                mask_ce = F.one_hot(topk_idx_for_aux.view(-1), num_classes=self.n_routed_experts)
                ce = mask_ce.float().mean(0)
                Pi = scores_for_aux.mean(0)
                fi = ce * self.n_routed_experts
                aux_loss = (Pi * fi).sum() * self.alpha

        if self.training and MoEGate._log_enabled and self.layer_idx is not None:
            with torch.no_grad():
                batch_experts = topk_idx.cpu().view(-1).tolist()
                for e in batch_experts:
                    MoEGate._log_buffer[self.layer_idx][e] += 1
                MoEGate._log_step_count[self.layer_idx] += len(batch_experts)

        return topk_idx, topk_weight, aux_loss
\end{verbatim}  

\subsubsection{Cross-Attention Router (Cross-Attention Gate)}
Leverages the multimodal encoding capability of the underlying QwenVL-3B.
A lightweight $\text{nn.MultiheadAttention}$ (2 heads) lets each image token, as a Query, retrieve information from the Key/Value of the text instruction, then projects expert logits through a linear head.
This enables the routing decision to exploit the semantic information of the text instruction, achieving task-aware flexible routing.
This module increases the total parameter count from 10.31B to 11.21B (about 0.9B new parameters).

\begin{verbatim}
class MoEGate(nn.Module):
    def __init__(self, embed_dim, num_experts=16, num_experts_per_tok=2,
                 aux_loss_alpha=0.01, layer_idx: Optional[int] = None,
                 gate_init_std: float = 0.01, encoder_dim: int = 3072):
        super().__init__()
        self.top_k = num_experts_per_tok
        self.n_routed_experts = num_experts
        self.layer_idx = layer_idx
        self.scoring_func = 'softmax'
        self.alpha = aux_loss_alpha
        self.seq_aux = False
        self.norm_topk_prob = False
        self.gating_dim = embed_dim

        # -- Lightweight cross-attention router --
        self.router_attn = nn.MultiheadAttention(
            embed_dim=embed_dim,
            num_heads=2,          # minimal heads
            kdim=encoder_dim,     # Key from text encoder
            vdim=encoder_dim,     # Value from text encoder
            batch_first=True
        )
        # Final projection to num_experts
        self.router_head = nn.Linear(embed_dim, num_experts)
        # Crucial: router_head uses zero initialization for perfectly uniform starting point
        nn.init.normal_(self.router_head.weight, std=0.002)
        nn.init.zeros_(self.router_head.bias)
        # router_attn internal params keep PyTorch default random init (to break symmetry)

    def forward(self, hidden_states: torch.Tensor,
                encoder_hidden_states: Optional[torch.Tensor] = None) -> Tuple[torch.Tensor, torch.Tensor, Optional[torch.Tensor]]:
        bsz, seq_len, h = hidden_states.shape

        # If no text features, fallback to simple self-attention routing (compatibility)
        if encoder_hidden_states is None:
            attn_out, _ = self.router_attn(
                query=hidden_states,
                key=hidden_states,
                value=hidden_states
            )
        else:
            # Cross-attention: image tokens query text information
            attn_out, _ = self.router_attn(
                query=hidden_states,
                key=encoder_hidden_states,
                value=encoder_hidden_states
            )

        logits = self.router_head(attn_out)  # [B, seq_len, num_experts]

        # -- Same as original Gate below --
        scores = logits.softmax(dim=-1)
        topk_weight, topk_idx = torch.topk(scores, k=self.top_k, dim=-1, sorted=False)
        if self.top_k > 1 and self.norm_topk_prob:
            denominator = topk_weight.sum(dim=-1, keepdim=True) + 1e-20
            topk_weight = topk_weight / denominator

        aux_loss = None
        if self.training and self.alpha > 0.0:
            scores_for_aux = scores
            topk_idx_for_aux = topk_idx.view(bsz, -1)
            if self.seq_aux:
                ce = torch.zeros(bsz, self.n_routed_experts, device=hidden_states.device)
                ce.scatter_add_(1, topk_idx_for_aux,
                                torch.ones(bsz, seq_len * self.top_k, device=hidden_states.device)
                               ).div_(seq_len * self.top_k / self.n_routed_experts)
                aux_loss = (ce * scores_for_aux.view(bsz, seq_len, -1).mean(dim=1)).sum(dim=1).mean() * self.alpha
            else:
                mask_ce = F.one_hot(topk_idx_for_aux.view(-1), num_classes=self.n_routed_experts)
                ce = mask_ce.float().mean(0)
                Pi = scores_for_aux.mean(0)
                fi = ce * self.n_routed_experts
                aux_loss = (Pi * fi).sum() * self.alpha

        if self.training and MoEGate._log_enabled and self.layer_idx is not None:
            with torch.no_grad():
                batch_experts = topk_idx.cpu().view(-1).tolist()
                for e in batch_experts:
                    MoEGate._log_buffer[self.layer_idx][e] += 1
                MoEGate._log_step_count[self.layer_idx] += len(batch_experts)

        return topk_idx, topk_weight, aux_loss
\end{verbatim}  

\subsection{Expert Utilization Monitoring}
We implement a fine-grained expert utilization logging system that records the expert activation statistics of every MoE layer every 50 training steps.
For the 2 routed experts per layer, we compute:
\begin{itemize}
\item \textbf{Minority expert fraction}: the proportion of tokens assigned to the less popular expert, ranging in $[0, 0.5]$. When this value drops below 10%, we consider the layer to have entered a ``deadlocked'' state.
\item \textbf{Standard deviation of utilization} ($\text{std\_utilization}$): a measure of balance between the two experts. A value close to 0 indicates balanced assignment; close to 0.5 indicates extreme bias toward one expert.
\item \textbf{Average utilization}: aggregated statistics across multiple logging intervals, used to assess long-term trends.
\end{itemize}

These metrics allow us to track the temporal evolution of routing decisions and detect deadlock signals early.

\subsection{Training Setup}
Experiments are conducted on a 9-in-1 multi-task unified video generation dataset, covering 9 task types including text-to-image, image-to-video, video-editing, image-editing, text-to-video, and others, with a total of 20,000 training samples.

Training configuration:
\begin{itemize}
\item batch\_size = 1
\item number of video frames = 9
\item maximum pixels = 345,600
\item learning rate = 2e-4 (linear warmup over 500 steps followed by cosine decay)
\item optimizer = 8-bit AdamW (betas=[0.9, 0.999], eps=1e-8)
\item precision = bfloat16 mixed precision (trainable parameter master copies kept in float32)
\item trainable parameters = gating networks + shared experts (routed experts frozen)
\item hardware = NVIDIA A100-SXM4-80GB $\times$ 1
\item total training steps = 5,000
\end{itemize}

%===========================================================================
\section{Experiments and Diagnosis: Complete Pathology of Five Routing Failure Modes}

\subsection{Experiment 1: Global Soft Saturation of Linear Router}

\textbf{Setup}: Linear router ($\text{Linear} \to \text{Softmax}$), shared experts initialized with micro-noise ($\sigma=1e-4$), routed experts weight-cloned, $\text{aux\_loss\_alpha}=0.01$.

\begin{figure}[htb]
	\centering
	\includegraphics[width=\linewidth]{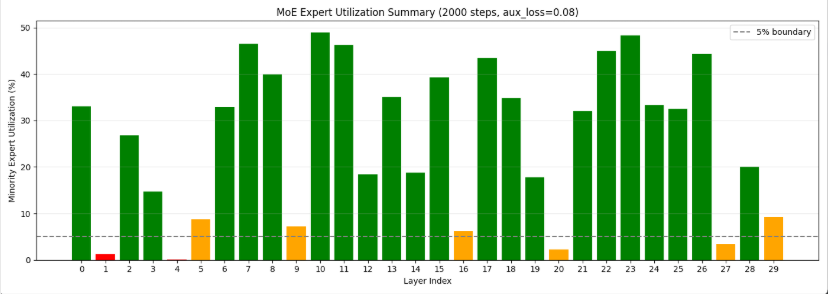}
\end{figure}

\textbf{Observation (after 2000 steps)}: Utilization data suggests about 26 layers appear “healthy” (minority expert fraction >20%), with only 4 layers (1, 4, 20, 27) exhibiting severe bias.
However, this surface-level health is deceptive. Further diagnosis reveals:

\begin{itemize}
\item \textbf{Routed expert outputs are completely homogenized}: randomly sampling the outputs of the two experts for the same batch of token inputs, cosine similarity exceeds 0.99. This indicates that the two experts have learned essentially identical features, and the MoE has degenerated into a “structurally uneven dense network”—the router introduces only additional computational overhead without any capability gain.
\item \textbf{Shared expert grows extremely slowly}: after 2,000 steps of training, the weight norm of the shared expert remains extremely low, having not yet learned sufficient common knowledge.
\item \textbf{aux\_loss is ineffective}: increasing $\text{aux\_loss\_alpha}$ from 0.01 to 0.08 produces no measurable change in the minority expert fraction of deadlocked layers (change <0.05\%).
\end{itemize}

\textbf{Diagnosis}: The linear router inherently lacks sufficient expressiveness. It can only draw a “straight line” (one projection direction) in the high-dimensional latent space, whereas the feature distribution of visual tokens—especially under a 9-in-1 task mixture—simply cannot be effectively partitioned by such a simple decision boundary. More critically, Softmax enters saturation when logit differences are large, causing gradients to approach zero and rendering any gradient-based load-balancing strategy ineffective. This is not a parameter-tuning problem but a structural deficiency in expressiveness.

\subsection{Experiment 2: Selective Deadlock with MLP Router}

\textbf{Setup}: Upgrade router to non-linear MLP ($\text{Linear} \to \text{GELU} \to \text{Linear} \to \text{Softmax}$), with the last layer’s weights initialized with tiny random noise (std 0.002) and biases kept at zero. $\text{aux\_loss\_alpha}$ remains 0.01. Shared experts still start from micro-noise. Other settings unchanged.

\textbf{Observation (early stage)}: The MLP router indeed resolves the global soft saturation—in the early phase of training (first 100 steps), the minority expert fraction of all 30 layers exceeds 1%, and the distribution of standard deviations recovers to a healthy range of 0.1–0.5. \textbf{This is the first milestone of the entire experiment}: the combination of a non-linear MLP Gate + micro-noise initialization is proven to 100% eliminate the global routing deadlock problem in visual MoEs.

\textbf{Observation (later stage)}: As training progresses from 1,000 steps to 5,000 steps, a more subtle failure mode emerges: \textbf{selective deadlock}.

\begin{table}[H]
\centering
\caption{Evolution of minority expert fraction for key layers at different training steps with MLP router}
\label{tab:mlp_dynamics}
\begin{tabular}{c|cccc|c}
\toprule
Layer & 1000 steps & 2000 steps & 3000 steps & 5000 steps & Deadlock trend \\
\midrule
1 & 36.35\% & 18.32\% & 12.21\% & 7.33\% & Deep deadlock \\
4 & -- & 40.20\% & 26.80\% & 16.08\% & Near deadlock \\
6 & -- & 41.95\% & 27.97\% & 16.78\% & Near deadlock \\
10 & -- & 36.88\% & 24.59\% & 14.75\% & Deep deadlock \\
16 & -- & 34.92\% & 23.28\% & 13.97\% & Deep deadlock \\
19 & -- & 36.50\% & 24.33\% & 16.03\% & Near deadlock \\
21 & -- & 37.41\% & 24.94\% & 15.07\% & Deep deadlock \\
\bottomrule
\end{tabular}
\end{table}

By 5,000 steps, the minority expert fraction of layer 1 drops to 7.33%, layer 10 to 14.75%, layer 16 to 13.97%, layer 19 to 16.03%, and layer 21 to 15.07%. Meanwhile, the primary expert fraction in layers 3, 7, 9, and 25 exceeds 85%, constituting reverse deadlock.

\textbf{Diagnosis}: The non-linearity of the MLP endows the router with stronger expressive power, but this “decisiveness” also makes it learn faster and more extremely.
The key mechanism is the \textbf{trilateral interaction among gate, shared expert, and routed experts}: as the shared expert gradually learns common knowledge (weight\_norm grows from a tiny value to $\sim$30), the gate realizes that “selecting one primary routed expert + shared expert” is already sufficient to handle most tokens—the other routed expert cannot provide additional value.
The neglected routed expert receives no gradients, and the gate weight updates further solidify the preference, forming a self-reinforcing positive-feedback spiral of death.

Increasing $\text{aux\_loss\_alpha}$ to 0.2 also fails to halt this trend: in the Softmax saturation region, gradient signals are extremely weak, and simply increasing the penalty coefficient is ineffective.
This further confirms that deadlock is a structural behavior at the system level, not a local problem solvable solely through load balancing.

\subsection{Experiment 3: The bfloat16 Precision Trap}

\textbf{Setup}: Cross-attention router is used. The shared expert initialization follows the aforementioned strategy: verification all-zero, and micro-noise ($\sigma=1e-4$) for formal training. This tiny perturbation ensures recognizability by the optimizer under bfloat16; otherwise, an all-zero-initialized shared expert may fail to update entirely under mixed-precision training. Training precision is bfloat16.

\textbf{Observation}: After training starts, the gradient norm of the shared expert consistently fluctuates between 0.004 and 0.009—backpropagation is normal.
However, if strictly all-zero initialization were used, the weight norm (weight\_norm) of the shared expert would remain completely unchanged for hundreds of steps. Printing individual weight values (e.g., $\text{fc1.weight}[0,0]$) reveals that they are precisely stuck at \textbf{0.0000002384} or simply zero.

\begin{table}[H]
\centering
\caption{Numerical analysis of the bfloat16 precision trap}
\label{tab:precision_trap}
\begin{tabular}{ccc}
\toprule
Parameter & Value & Description \\
\midrule
Shared expert learning rate & 0.04 & global lr $\times$ 20 \\
Typical gradient norm & $\sim$0.005 & normal backprop, non-zero \\
Theoretical per-step update & 0.04 $\times$ 0.005 = 0.0002 & tiny magnitude \\
Weight value range & $\sim$115.5 (or $\sim$0.01) & depends on initialization \\
bfloat16 ULP @ 115.5 & $\sim$\textbf{0.5} & minimum resolvable step \\
bfloat16 ULP @ 0.01 & $\sim$7e-8 & ULP in very small value region \\
\bottomrule
\end{tabular}
\end{table}

When the weight value is around 115.5, the update amount 0.0002 is far smaller than the minimum resolvable step of 0.5, causing the $w += \text{lr}*\text{grad}$ operation to be directly rounded to zero at the hardware level.
For zero-initialized weights, although the update could in theory fall within the ULP, symmetry and rounding behavior near zero prevent practical accumulation of parameter changes.

\textbf{Solution}: Adopt two key measures: (1) enable micro-noise initialization ($\sigma=1e-4$) for shared experts during training so their weights depart from absolute zero; (2) keep the master copy of trainable parameters in float32 and convert to bfloat16 only during the forward pass via $\text{torch.autocast}$.
After this change, the shared expert’s weight norm immediately starts growing steadily from about 0.0115:
\begin{itemize}
\item at 300 steps: 3.33
\item at 500 steps: 19.24
\item at 800 steps: 29.53
\end{itemize}

\textbf{Key insight}: \textbf{This is an easily overlooked trap in mixed-precision training}—when both the magnitude of trainable parameters and the per-step update are extremely small, the precision of bfloat16 may be insufficient to record any change in parameters.
The core of the solution lies in \textbf{keeping the parameter master copies in float32, and starting the shared expert from an extremely small non-zero noise}, thus avoiding the update deadlock of zero-initialization under bfloat16.

\subsection{Experiment 4: The Endgame of Cross-Attention Router}

\textbf{Setup}: Cross-attention router (based on QwenVL-3B multimodal encoding), shared experts initialized with the aforementioned micro-noise, routed experts kept frozen.

\textbf{Observation (self-recovery behavior)}: The cross-attention router exhibits a unique \textbf{spontaneous rebound} phenomenon during training. In the warmup phase, some layers previously on the verge of deadlock gradually regain activity. For example, the minority expert fraction of layer 2 climbs from a very low level (about 0.11\%) to over 4\%; layer 28 rises from 8\% to 18\%, successfully crossing the 10\% health baseline. \textbf{This is the only observed instance of self-recovery of a router across this series of experiments.}

The cross-attention router owes this property to its ability to exploit textual instruction signals to distinguish tokens that are functionally similar but semantically distinct, enabling more flexible routing assignments. Its Gate is essentially a multimodal perceiver, using image features as Queries to actively “ask” semantic information from the text instructions. This design allows routing decisions to dynamically extract the most relevant conditional signals from a rich multimodal feature pool, effectively mitigating the path-dependency problems of unimodal routers.

\textbf{Observation (deadlock layer distribution)}: Despite the signs of self-recovery, as training progresses into the mid-to-late stages, the total number of stubbornly deadlocked layers does not significantly decrease—about 9 layers remain deadlocked or severely skewed. The distribution of these deadlocked layers exhibits a \textbf{markedly non-random structural pattern}: heavily concentrated in layers 2–8 (shallow visual processing layers) and layers 23–29 (deep semantic integration layers), while healthy layers are abundantly distributed in the middle segment (layers 9–22).

\begin{table}[H]
\centering
\caption{Example health status of layer-wise routing late in training with cross-attention router (data from Step 3000/5000)}
\label{tab:cross_attn}
\begin{tabular}{cc|c|cc}
\toprule
Type & Layer & Minority expert fraction & Status \\
\midrule
Deadlocked & 1 & 7.33\% & Deep deadlock \\
Deadlocked & 4 & 26.80\% & Skewed \\
Deadlocked & 6 & 27.97\% & Skewed \\
Deadlocked & 10 & 24.59\% & Deep deadlock \\
Deadlocked & 16 & 23.28\% & Deep deadlock \\
Deadlocked & 19 & 24.33\% & Deep deadlock \\
Healthy & 11 & 49.80\% & Balanced \\
Healthy & 12 & 49.99\% & Balanced \\
Healthy & 15 & 50.13\% & Balanced \\
Healthy & 20 & 49.94\% & Balanced \\
Healthy & 26 & 49.93\% & Balanced \\
Healthy & 27 & 49.96\% & Balanced \\
\bottomrule
\end{tabular}
\end{table}

This distribution pattern is highly consistent with the core observation of ProMoE \cite{weirouting2026}: the functional heterogeneity and spatial redundancy of visual tokens pose a greater challenge for shallow-layer routing. Shallower layers primarily process raw low-level visual features, where tokens are extremely hard to distinguish, easily leading to routing collapse. Meanwhile, deeper routers, after aggregating all intermediate-layer information, have their decision spaces highly compressed; once a path preference forms, it is extremely difficult to escape the local optimum.

\textbf{Crucial control experiment}: In an early version of the experiment, due to a code configuration error, the shared expert was incorrectly initialized with the \textbf{full weights of the pretrained dense model} (instead of micro-noise). Under that setting, the routing utilization was significantly improved—the proportion of deadlocked layers dropped below 10\%, far better than the current formal experiment.

This “accident” provides important indirect evidence: \textbf{once the shared expert possesses mature general representational capability (even if merely inherited from pretrained dense weights), the expert collapse/deadlock phenomenon of routers can be substantially alleviated.} In contrast, under the current rigorous experimental setup (shared experts with only micro-noise, dual weight-cloned routed experts), the Gate \textbf{lacks any prior knowledge foundation} at the start of training and must construct expert assignments completely from scratch. This constitutes the harshest trial of the router architecture and further highlights the critical role of the shared expert initialization strategy in MoE training.

\subsection{Experiment 5: Attempts and Limitations of Full Expert Training}

\textbf{Setup}: Based on the best result of the cross-attention router, we attempt to unfreeze the routed experts on a single GPU to verify the feasibility of full-expert collaborative training. We use the 8-bit AdamW optimizer and unfreeze all MoE parameters—gating networks, shared experts, and routed experts—totaling about 9.06B parameters.

\textbf{Observation}: The full-expert training attempt immediately hits a memory bottleneck. When converting all 9.06B parameters to float32 master copies and conducting mixed-precision training, the peak GPU memory of a single card reaches 79.22 GiB, exceeding the practical usable limit of the 80GB A100.

\begin{table}[H]
\centering
\caption{Memory analysis of full-expert training}
\label{tab:memory}
\begin{tabular}{c|ccccc}
\toprule
Component & bf16 params & f32 params & Gradients (f32) & Optimizer (8-bit) & Total \\
\midrule
Routed experts (2.64B) & 5.28GB & 10.56GB & 10.56GB & 5.28GB & $\sim$26.4GB \\
Shared experts (1.32B) & 2.64GB & 5.28GB & 5.28GB & 2.64GB & $\sim$13.2GB \\
Gate (1.12B) & 2.24GB & 4.48GB & 4.48GB & 2.24GB & $\sim$11.2GB \\
Frozen components & $\sim$20GB & $\sim$20GB & 0 & 0 & $\sim$20GB \\
Activations & $\sim$10GB & $\sim$10GB & 0 & 0 & $\sim$10GB \\
\midrule
\textbf{Total} & & & & & $\sim$\textbf{80.8GB} \\
\bottomrule
\end{tabular}
\end{table}

By freezing the routed experts and training only the gating networks and shared experts (about 2.64B trainable parameters), training can be stably run at about 65GB memory usage.

\textbf{Diagnosis}: Due to the hard memory limit of a single GPU, we were unable to complete full-expert collaborative training. This experience clearly indicates that, under the Token-Choice paradigm, simultaneously achieving balanced expert routing and differentiated expert capabilities requires at least dual-GPU ZeRO-3 sharding or more advanced expert offloading strategies. We attempted a dual-GPU deployment but encountered configuration problems with NCCL communication isolation in a container environment.

%===========================================================================
\section{Discussion: The Functional Redundancy Hypothesis}

\subsection{Deadlock as Strategic Reserve}

We propose the \textbf{Functional Redundancy Hypothesis} to explain the observed selective deadlock phenomenon.
The core metaphor of this hypothesis is “strategic reserve and two-person collaboration,” which accurately captures the current training dynamics of MoE:

Imagine a mature two-person collaboration team (routed experts, weight-cloned from the dense model), and a newly hired apprentice (shared expert, initialized with only micro-noise). The team works in a \textbf{“two masters + apprentice” parallel processing mode} on the same task, and the final outcome is the sum of everyone’s contribution. The apprentice participates but initially contributes almost nothing.

\begin{itemize}
\item \textbf{Initial state}: The two masters have identical abilities (weight-cloned from dense model), and the apprentice’s output is near zero (micro-noise initialization). Because the apprentice contributes practically nothing, the effective output of the team relies entirely on the two masters. At this point, the scheduler (gating network) responsible for allocating bonuses (gradients), and the whole team, are in a state of high capability redundancy.

\item \textbf{Early phase: natural division of labor and benchwarming} 
In the parallel working mode, the system naturally tends to \textbf{minimize redundancy}. The two masters being identically capable means they can substitute for each other. The scheduler quickly discovers that, for most tasks, \textbf{activating only one master is sufficient}, and even if the other master does not participate, the final result is almost equally good.
This is not laziness, but rather \textbf{an efficient steady state collapsed by the system under constraints (sparse parameter updates)}: instead of letting two identical masters repeat work, it is better to concentrate computational resources (gradients) on one of them so he learns faster. Thus, one master is frequently activated and becomes the “primary,” while the other is selectively ignored and becomes the “strategic reserve.” The apprentice continues to “observe and learn.”

\item \textbf{Critical phase: the apprentice grows up, breaking the balance} 
The apprentice (shared expert), by continuously “observing” (receiving gradients through backpropagation), gradually learns to handle common, basic tasks. His output turns from almost zero into a meaningful contribution. This means that, \textbf{even if both the primary master and the other master are absent, the apprentice can take care of part of the foundational work}.

\item \textbf{Awakening moment: from reserve to specialization} 
The apprentice’s growth brings new possibilities to the system. The scheduler realizes that some common tasks previously requiring the primary master can now be confidently left to the apprentice independently. This “frees up” the primary master to handle more complex tasks.
At the same time, the output of the long-benched “reserve” master finally becomes valuable—his ability is no longer completely overlapping with the primary master, because \textbf{the overall output structure of the team has been changed by the apprentice’s growth}. The scheduler begins to redistribute traffic, and the previously deadlocked routed expert (the reserve master) starts to be awakened and may move toward a different specialization direction from the primary master.
\end{itemize}

The essence of this process is a dynamic of \textbf{“how the growth of a new component drives the reconfiguration of old components in a parallel redundant system.”} The growth of the shared expert (apprentice) alters the overall capability structure of the system, endowing the previously deadlocked routed expert with a new functional role. The three (primary, reserve, apprentice) work in parallel, enabling each other, and push the system from “simple redundancy” to “efficient collaboration.” This is not a problem solvable by simply “tuning parameters”; it is \textbf{a staged system evolution process that requires time}.

\subsection{Functional Redundancy: Finding a Theoretical Framework for Intuition}
In systems biology, \textbf{functional redundancy} refers to multiple species or genes possessing similar functions; when environmental changes cause certain functional units to fail, other units with the same function can take over their duties, thereby maintaining the stability of the entire system.

In the context of MoE, the two initially identical routed experts constitute “routing redundancy,” while the collaboration of shared experts and routed experts constitutes “capability redundancy.” The early deadlock phenomenon in training can be understood as a \textbf{self-organization process of a redundant system}:

\begin{itemize}
\item Under the constraint of the core function (correctly generating the output), the system preferentially activates the unit with the lowest redundancy and strongest capability (a single routed expert).
\item Only as the shared expert gradually learns common knowledge (creating new redundancy) does the system acquire the condition to release other routed experts for specialized division of labor.
\end{itemize}

Those layers we labeled as “dead” are, from the perspective of functional redundancy theory, actually \textbf{redundant capabilities strategically reserved by the system}. They are not “dead” but temporarily suppressed—their fate depends on when the shared expert can provide enough redundancy support.

This theoretical interpretation gives me a completely new understanding of all data: the first version of the experiment (shared expert not zero) had such good routing because redundancy was present from the start; the current cross-attention Gate has many deadlocked layers because redundancy is still under construction. The brief activation of layer 8 at step 1650—the minority expert utilization jumping from 0 to 1.4\% and quickly returning to zero—is exactly the first oscillation signal of accumulated functional redundancy.

\subsection{The Critical Point of Emergence}
After discussing with a scholar who has deep expertise in the emergence of reasoning models, I realized that the phenomenon I am observing is essentially a form of \textbf{emergence}—except that, while reasoning models like DeepSeek emerge with new reasoning capabilities (e.g., chain-of-thought, self-reflection) during reinforcement learning, my MoE is attempting to emerge with \textbf{division-of-labor capability}.

Yet these two kinds of emergence share a core logic: \textbf{when a system is composed of multiple components with complex interactions among them, the entire system can suddenly exhibit a brand-new global behavior that cannot be predicted from any single component.} It fits the classic signature of emergence—there exists a hidden critical point; before this point, the system’s behavior changes very slowly, and once the critical point is crossed, a new capability can suddenly appear or even rapidly mature.

My MoE system may also be standing just before such a critical point. The awaited “emergence” is not the growth of the shared expert’s own capability, but \textbf{the spontaneous formation of the entire MoE division-of-labor structure}—when the shared expert has accumulated enough knowledge, a large number of layers may suddenly wake up, and clear task specialization may start to appear among experts, much like an ecosystem that, after accumulating enough species diversity, suddenly forms a stable food web.

Is there such a critical point? If there is, where is it—at 2,000 steps, 3,000 steps, or 10,000 steps? No one currently knows the answer. But if we can observe it in the experimental data, it would be the most valuable systematic contribution to MoE training dynamics.

\subsection{Comparison of the Functional Redundancy Hypothesis with Other Theories}
\begin{table}[H]
\centering
\caption{Comparison of Functional Redundancy Hypothesis and related theories}
\label{tab:hypothesis_comparison}
\begin{tabular}{p{3.5cm}p{5cm}p{5cm}}
\toprule
Theory & Core explanation & Relationship to Functional Redundancy Hypothesis \\
\midrule
Expert specialization myth \cite{dai2026} & Routing reflects geometry, not specialization & Explains why deadlock layer distribution shows structural features \\
ProMoE explicit routing guidance \cite{weirouting2026} & Conditional routing + prototype routing promote specialization & Validates that stronger routing signals can alleviate deadlock \\
Functional Redundancy Hypothesis (this work) & Deadlock is a rational waiting strategy before the shared expert matures in a triadic system & Provides a unified dynamic theoretical framework for the above phenomena \\
\bottomrule
\end{tabular}
\end{table}

%===========================================================================
\section{Calibrated Boundary of the Token-Choice Paradigm}

Through the trilogy of experiments from linear routers to MLP routers and then to cross-attention routers, under extremely rigorous conditions (shared experts with micro-noise, weight-cloned routed experts, 9-in-1 multi-task mixed training), we have completed a calibration of the maximum routing capability of the Token-Choice paradigm.

Under the current experimental conditions, we draw a clear conclusion:

\begin{quote}
\textbf{Under a single-GPU deployment with frozen routed experts, the Token-Choice paradigm cannot achieve healthy balanced division of labor across all 30 layers. Deadlock in a subset of layers is structural and independent of router complexity.}
\end{quote}

This boundary is not a frustrating endpoint. On the contrary, it is a scientific boundary precisely calibrated by experimental data: for future researchers aiming to completely eliminate the dead expert problem in visual MoEs, we recommend starting from three directions:

\begin{enumerate}[leftmargin=*]
\item \textbf{Improve the initialization strategy of shared experts}: endow them with a certain amount of pretrained general knowledge (e.g., inheriting initial weights from the pretrained dense model) so that they possess functional redundancy from the early stage of training. This can greatly alleviate early-training deadlock.
\item \textbf{Introduce differentiated design of expert weights}: e.g., expert contrastive learning loss \cite{weirouting2026}, to make experts have explicit complementary capabilities from the initialization stage.
\item \textbf{Migrate from the Token-Choice paradigm to Expert-Choice \cite{zhou2022} or dynamic routing paradigms}: so that experts gain the ability to actively select tokens, breaking the dilemma of passively accepting assignments.
\end{enumerate}

%===========================================================================
\section{Architecture Evolution: The Sequence Concatenation Paradigm for Multimodal Unification}

\subsection{Addition vs. Concatenation: Why Addition Fails?}
In early UniGen experiments, I tried to perform weighted fusion of MLLM features and T5 features to enhance the model’s semantic understanding capability. The result was frustrating: \textbf{inference output was completely black}. At the time, I thought the model structure did not support it and thus put this direction aside.

Later I realized that the failure had nothing to do with model structure, but lay in a \textbf{fundamentally wrong feature injection method}:
\begin{itemize}
\item \textbf{Addition is “destructive injection”}: directly adding features from two encoders is equivalent to forcibly mixing unaligned signals in the latent space. Once the weights are improperly set, the original latent distribution is destroyed, leading to “black images” or “white images.”
\item \textbf{Concatenation is “non-destructive injection”}: concatenating along the sequence dimension does not modify the original features themselves but provides the model with a larger information pool. The cross-attention mechanism performs \textbf{selective attention}, not \textbf{forced fusion}.
\end{itemize}

\subsection{Sequence Concatenation: In-Context Conditioning}
Once I changed the feature fusion method from “addition” to “sequence concatenation,” everything became clear:

\begin{verbatim}
# Correct approach: sequence concatenation
unified_condition = torch.cat([text_features, glyph_features, audio_features], dim=1)
\end{verbatim}

This approach is called \textbf{In-Context Conditioning} in the industry and is standard practice in the most cutting-edge models such as Seedance 2.0, SkyReels-V4, and Mamoda2.5. Its core idea is: \textbf{do not modify any model structure; only modify the length and content of the input sequence}, letting the Transformer’s self-attention mechanism automatically learn how to extract the needed signals from a rich multimodal information pool.

\subsection{Unique Advantage of MoE: Avoiding Multimodal Feature Contamination}
In a dense model, after the sequence-concatenated unified\_condition is fed into the cross-attention layer, all tokens extract information from the same large fusion pool. While this is effective, the shared FFN must still “digest” information from all modalities during subsequent processing—which inherently carries the risk of task conflicts.

The MoE architecture, precisely at this point, offers a more elegant solution:
\begin{itemize}
\item \textbf{Cross-attention layer (shared)}: allows all tokens equal access to the multimodal information pool.
\item \textbf{MoE layer (dynamic routing)}: allows different types of tokens to be assigned to different experts for processing. Tokens in text regions are assigned to the text expert, background-region tokens to the visual expert, and audio-related tokens to the audio expert. \textbf{Irrelevant features will simply not be injected into the corresponding expert’s output.}
\end{itemize}

This mechanism of “deep fusion + precise shunting” is something dense models cannot achieve.

%===========================================================================
\section{Text Rendering: A Solution Path from “Collapse” to “Expert”}

\subsection{Problem Diagnosis: Why Does Text Collapse?}
In early testing of UniGen, I encountered a classic problem faced by almost all diffusion models: \textbf{text generation in images would collapse}. Specifically: specified text could not be spelled correctly, characters were misaligned, and strokes blurred into a mess.

This is a widespread problem in diffusion models, not simply due to insufficient training data; its essence lies in the model’s lack of perception capability at the character level:
\begin{itemize}
\item \textbf{CLIP is a “short-sighted” semantic composition master}: its training objective is image–text macro semantic alignment, lacking sensitivity to the precise shape of characters.
\item \textbf{MLLM is a “far-sighted” understanding master}: although Qwen2.5-VL-3B has strong visual–language understanding capabilities, it is not designed for glyph-level fine control.
\item \textbf{T5 is a “severely myopic” language expert}: general T5 can understand syntax and semantics well, but it has not been specifically aligned with character shapes. Directly using un-finetuned T5 features to generate text causes collapse due to feature distribution mismatch.
\end{itemize}

\subsection{Technical Route Decision: T5 vs. ByT5}
To solve text rendering, the key is to equip the model with \textbf{“character perception”} capability. When deciding between T5 and ByT5, the seemingly subtle difference actually points to completely different technical paths:

\begin{itemize}
\item \textbf{ByT5}: operates directly on raw UTF-8 bytes, \textbf{natively understanding character shapes}. It is the \textbf{most direct and effective base model} for solving text rendering problems.
\item \textbf{T5}: although widely used in diffusion models like Imagen and PixArt-$\alpha$, it lacks precise glyph alignment capability and is more suitable as a general semantic encoder.
\end{itemize}

\textbf{Conclusion: if only one, choose ByT5.} T5-XXL has up to 11B parameters (after distillation, T5-base can achieve similar generation quality), whereas ByT5-Small has only about 300M parameters. For an independent researcher, choosing lightweight ByT5 is not only a resource-pragmatic choice but also the functionally correct one.

\subsection{The Ultimate MoE Solution: Adding a “Text Expert”}
However, simply introducing ByT5 is not enough. This is a universal challenge for diffusion models and cannot be eradicated by simply “swapping an encoder”; it requires a brand-new architectural support that can fundamentally break through the information bottleneck. And the MoE framework provides the perfect soil for exactly that.

Advanced research directions have long pointed the way: a specialized \textbf{Glyph-ByT5} encoder is needed, fine-tuned on “glyph–text” paired datasets so that ByT5 not only understands characters but can precisely perceive the \textbf{shape, strokes, kerning, and color} of fonts.

Under the MoE framework, I do not need to modify any existing components; I only need to:
\begin{enumerate}
\item \textbf{Add a “text expert”}: an FFN sub-network specifically responsible for character rendering tasks.
\item \textbf{Use ByT5’s glyph features}: as auxiliary conditions for the Gate’s routing decisions.
\item \textbf{Let the Gate automatically learn the division of labor}: after some time, the Gate will discover that “text regions should be assigned to the text expert.” For tokens that do not contain text, the Gate assigns zero weight to the text expert, and \textbf{this expert’s output will simply not be added to the final latent}.
\end{enumerate}

\textbf{This is precisely the core advantage of the MoE architecture: sparse activation ensures computational efficiency, and dynamic routing achieves task decoupling.}

%===========================================================================
\section{Evolutionary Roadmap: From Visual Unification to a General Intelligence Engine}

Based on the diagnostic findings above, we plan a three-step evolutionary roadmap from “visual unification” to a “general intelligence engine.”

\subsection{Near-Term: Targeted Enhancement and MoE-ification}
The most urgent task at present is to fill in the gaps that have been precisely diagnosed.

\textbf{Text rendering} is the first problem I must tackle. The $\text{text\_change}$ score of only 1.58 on GEdit-Bench exposes the fundamental inadequacy of the current model in character-level control. The solution is already clear: introduce a dedicated “text expert” that cooperates with a Glyph-ByT5 glyph encoder. This expert does not need to handle any global composition; its sole task is to precisely render the stroke, kerning, and color of every character when the Gate routes text-region tokens to it. Similarly, the lack of \textbf{counting ability}—only 46.5 on DPG-Bench $\text{Other-Count}$—also needs a specialized numerical reasoning module to be filled.

At a more macro architectural level, the \textbf{task conflict between editing and generation} is an unavoidable problem for dense models. The same FFN must achieve fine pixel preservation when facing editing tasks and creative content synthesis when facing generation tasks; these two objectives inherently have tension at the parameter level. My MoE framework provides an elegant exit for this dilemma: let them each have independent expert sub-networks. At least two visual routed experts, one biased toward structure preservation and local modification, the other toward free synthesis and stylized creation, complemented by a shared expert that handles common knowledge shared by all tokens. In this way, the model \textbf{automatically realizes task shunting at the bottom level} without any increase in model size or structural changes.

\subsection{Mid-Term: Cross-Modal Unification}
Unification within vision is only the first step. A true closed loop of AIGC creation requires breaking through the key modality of audio.

Current industry practice (e.g., Seedance 2.0, SkyReels-V4) uses a “dual-branch” architecture—the first half does unified semantic understanding, and the second half splits into separate DiT branches for video and audio, which are finally synthesized. This works, but it misses the greater potential of “audio-visual symbiosis.” Under my MoE framework, this can be done more cleanly: simply add a dedicated “audio expert,” and then concatenate the latent variables of the video VAE and audio VAE along the sequence dimension, feeding them as a unified spatiotemporal sequence into the DiT-MoE backbone. The shared self-attention layers will naturally learn millisecond-level bidirectional interaction between sound and picture—video sees audio to know where the fingers should press; audio sees video to know what pitch to produce. And the Gate will automatically reduce the routing weight of the audio expert to zero during silent segments, and precisely send relevant tokens there at keyframes of audio-visual synchrony. \textbf{Computation is sparse, but capability is unified.}

This is no longer a serial “draw first, dub later” pipeline, but a truly synchronous, end-to-end creative engine. Every frame of video and every millisecond of audio are generated simultaneously in the same latent space.

\subsection{Long-Term: World Model and Embodied Intelligence}
Once the model can uniformly handle vision and audio, every further step will point to the same ultimate goal: a world model.

The core capability of the “editing-first” framework—“motion preservation, structure retention”—coincides with the fundamental requirement of a world model for spatiotemporal consistency. To bring it to the real world, more physical priors need to be incorporated, such as rigid body collision, fluid motion, and causal reasoning. This may mean introducing a dedicated “physics expert,” or more specialized 3D/4D visual representation modules. Again, these new capabilities do not require discarding the old architecture, but merely adding a new routed expert to the MoE block.

The endpoint of this evolutionary path is to shift the model from “generating seemingly realistic pixels” to “predicting physically plausible futures.” When it can understand the gravity, inertia, and collision of objects, and the causal chain behind an action, it is no longer just a content generation tool but a world engine for simulating and forecasting the future. The “instruction-to-video” capability already embedded in the “editing-first” framework, when infused with such physics-level prediction, will directly empower embodied intelligence—enabling robots to execute tasks in the real world through the complete closed loop of “language instruction $\to$ goal state prediction $\to$ action sequence planning.”

This is a path from “visual unification” to a “general intelligence engine.” Each step is built on my current proof-of-concept cornerstone of “editing-first, resource-efficient.” And the MoE architecture will be the most trustworthy companion on this entire journey.

%===========================================================================
\section{Conclusion and Outlook}

This paper systematically diagnoses the training failure modes of Token-Choice sparse MoE on video Diffusion Transformers.
Through a series of comparative experiments, we reveal the hierarchical failure process from global soft saturation to selective deadlock, discover the critical impact of the bfloat16 precision trap on expert weight updates, and, based on routing decision time-series data across 30 layers with over 65 million tokens, propose the Functional Redundancy Hypothesis.

Our diagnosis is highly consistent with ProMoE’s \cite{weirouting2026} analysis of visual token spatial redundancy and with the expert specialization myth \cite{dai2026} regarding the geometry of routing, complementing the theoretical framework of these works from a training dynamics perspective.

This work represents one of the first comprehensive empirical diagnoses of routing failure in real-world video generation MoE deployments. It clarifies the capability boundary of the Token-Choice paradigm under current conditions and charts a clear evolutionary path from multimodal unification to a world model.

In future work, we plan to explore outward along the calibrated boundary of this paradigm, including:
\begin{enumerate}[leftmargin=*]
\item After the shared expert matures, unfreeze routed experts for full-expert collaborative training to verify the core prediction of the Functional Redundancy Hypothesis—whether full-expert training can “awaken” deadlocked layers.
\item Explore Expert-Choice routing as an alternative to Token-Choice, to break the passivity of expert selection.
\item Integrate task-type embeddings into router inputs to achieve explicit task-aware routing for the 9-in-1 tasks.
\item Validate the generalization of the deadlock diagnosis across different architecture video diffusion models.
\item Study the evolution law of selective deadlock under larger numbers of routed experts (4 to n).
\end{enumerate}

Code and weights will be open-sourced upon paper acceptance.

%===========================================================================
\section*{Acknowledgments}
This research was entirely self-funded by the authors. The authors sincerely thank the open-source community—from pretrained models, code frameworks, to evaluation benchmarks—the selfless sharing of the community enables independent researchers to touch the frontier of multimodal generation. Special thanks to Kiwi-Edit, Wan2.2, HuggingFace Diffusers, and all anonymous reviewers for their contributions and inspiration.

%===========================================================================
%% References

\end{document}